\documentclass[journal]{IEEEtran}
\usepackage{framed}
\usepackage[ruled]{algorithm2e}
\usepackage{amssymb}
\usepackage{latexsym}
\usepackage{epsfig}
\usepackage{float}
\usepackage{graphicx}
\usepackage{booktabs,multirow, multicol}
\usepackage{cite}
\usepackage{epsfig}
\usepackage{caption}
 \usepackage{rotating}
\usepackage{epstopdf}

\usepackage{tabularx}
\usepackage{url}
\usepackage{xcolor}
\definecolor{newcolor}{rgb}{.8,.349,.1}

\graphicspath{{pics/}}
\begin{document}

\graphicspath{{pics/}}
\title{Recognition of Activities from Eye Gaze and Egocentric Video}

\author{Anjith~George,~\IEEEmembership{Member,~IEEE,}
        and~Aurobinda~Routray,~\IEEEmembership{Member,~IEEE}
\thanks{ Thanks!}}

\maketitle

\begin{abstract}

This paper presents a framework for recognition of human activity from egocentric video and eye tracking data obtained from a head-mounted eye tracker.  Three channels of information such as eye movement, ego-motion, and visual features are combined for the classification of activities. Image features were extracted using a pre-trained convolutional neural network. Eye and ego-motion are quantized, and the windowed histograms are used as the features. The combination of features obtains better accuracy for activity classification as compared to individual features.

\end{abstract}


\section{Introduction}
Activity recognition from videos is an important topic in computer vision community. Recognition of actions has several applications in many areas such as human-computer interaction (HCI), robotics, surveillance, image and video retrieval. Most of the literature in this field deals with action recognition from video streams captured by a camera which may be situated far away from the subjects (third person view) \cite{poppe2010survey},\cite{weinland2011survey}, \cite{turaga2008machine}.

 Recently with the proliferation of wearable devices, there has been an upsurge in research in the field of activity recognition from wearable devices.  Recent works in egocentric video-based (first person view) activity recognition \cite{fathi2011understanding},\cite{pirsiavash2012detecting},\cite{yan2015egocentric} has shown great promise in providing insights into various activities. The egocentric video gives direct information regarding user's environment. Head-mounted eye trackers can provide gaze locations and head movements along with the ego-centric video.  

Nowadays a lot of virtual and augmented reality (VR and AR) devices are coming up in the consumer market such as Oculus Rift, Hololens, Google Glass \cite{starner2013project}, etc. They hold the potential to augment human capabilities. Eye tracking \cite{george2016fast} and egocentric video could give important cues about the user's point of attention and actions. Usage of visual features along with the eye movement behavior as observed through eye tracking can lead to the understanding of activities and cognitive processes.Identification of human emotions \cite{happy2012real} and cognitive states \cite{dasgupta2013vision}, \cite{sengupta2017multimodal} can lead to intelligent interaction modalities. Eye movements have shown to contain information useful for biometric authentication \cite{george2016score}.  Identification of human actions and intentions in real-time could result in human-machine systems which are more natural and `pro-active' .

\begin{figure}[t]
\begin{center}
\includegraphics[width=1\linewidth]{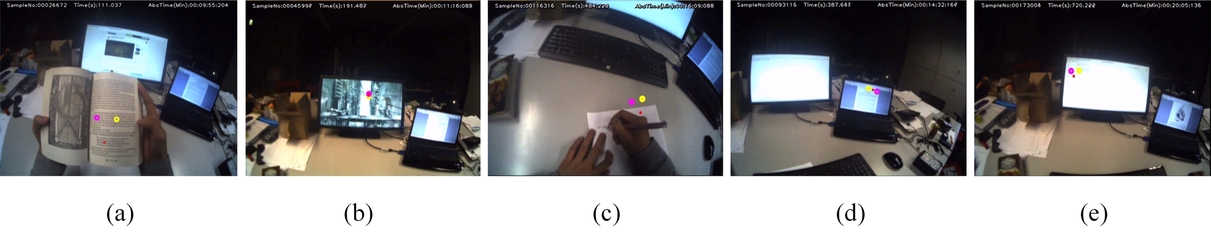}
\end{center}
\vspace{-1.5em}
\caption{The activity classes considered in the work, a) Read, b) Watching Video, c) Write, d) Copying text, and e) Browsing.}
\label{fig:samples}
\end{figure}

In this chapter, a framework for activity classification using egocentric information obtained from a head-mounted eye tracker is presented. Three channels of information, namely, eye movement patterns, ego-motion patterns and visual features as observed through the camera, are used for activity classification.  We consider activities performed in office environments which are difficult to classify by other modalities alone. Combining all these modalities can improve the accuracy of classification. The activity classes used in this work are shown in Fig. \ref{fig:samples}.



\section{Related works}


An excellent review of recent works in egocentric activity recognition can be found in \cite{nguyen2016recognition}.  Some of the recent works related to activity recognition from eye gaze are described here.

Bulling \textit{et al}.\cite{bulling2011eye} presented an activity recognition scheme based on eye movement parameters obtained using Electro Oculogram (EOG). They extracted a large number of features from fixations, saccades, and blinks. A feature selection approach was used to select the best features for activity classification. They considered five activities performed in the office environment, along with a null class. A support vector machine based classification was adopted for recognizing the activities. This work paved the way for further investigations using eye gaze where activity recognition using other modalities are difficult. Hipiny and Mayol-Cuevas \cite{hipiny2012recognising} presented an activity classification scheme using the gaze data. They represented each activity as a record of fixation locations. A Bag of words based weighted voting scheme, along with the Bhattacharya distance between templates and samples were used for classification. Ogaki \textit{et al}.\cite{ogaki2012coupling} presented an approach for egocentric activity recognition by fusing eye movement and ego-motion features.  They estimated ego-motion from the global optical flow computed from the “outward looking” camera. The eye tracking data was obtained from a head-mounted eye tracker. Both eye motion and ego-motion parameters were encoded to a string sequence using the motion pattern. The ‘N-gram’ statistics, computed over a sliding window, was used as a feature for classification.  From the experiments, they demonstrated that the combination of features improves the accuracy compared to eye movement features alone. George \textit{et al}. \cite{george2016real} presented an approach for gaze direction classification using Convolutional Neural Network. The direction of eyes obtained can provide important cues for activity recognition.

 Li \textit{et al}.\cite{li2015delving} presented a novel scheme for combining different modalities of information for egocentric action recognition. From the egocentric video, they extracted dense trajectories and a set of local descriptors across the trajectories. The features included motion binary histograms along $x$ and $y$ directions, histogram of flow, histogram of gradients and Lab color histogram. They computed these features within a grid, and the features were then concatenated. Egocentric features such as head motion and hand manipulation point were also extracted. They encoded the features using Improved Fisher Vector (IFV). Finally, the IFVs of different features were concatenated as a representation of the video. Support vector machine (SVM) was used for classification. However, they did not use eye movement patterns in their framework. Fathi \textit{et al}.\cite{fathi2012learning} demonstrated the relation between the task being performed and the locations of visual attention.  They showed that the information regarding hand-eye coordination could be beneficial in two different scenarios, predicting the probable gaze sequence given an action and predicting the likely action given the gaze sequence. Shiga \textit{et al}.\cite{shiga2014daily} proposed a method for egocentric activity recognition by combining eye motion and visual features.  The eye movement feature extraction scheme was similar to the method used in \cite{bulling2011eye}. They used ‘N-gram’ statistics computed over sliding windows. The visual features were obtained by selecting a patch around the gaze location and extracting local features using SIFT-PCA and dense sampling. A Bag of words approach was used for the classification. They trained separate multi-class SVMs for visual and eye movement features, and score fusion methodology was adopted for the final activity classification. Yan \textit{et al}.\cite{yan2015egocentric} proposed a multi-task clustering approach for egocentric activity classification.  They proposed two different algorithms for activity classification in unsupervised settings. Kunze \textit{et al}.\cite{kunze2013activity} provided a description of possibilities of eye tracking in various use cases such as detection of fatigue and reading. Data from mobile eye trackers can be utilized for the analysis of reading habits, type of document read, reading speed comprehension level and identifying alertness levels.




While there are many approaches for activity classification in egocentric videos, classification in indoor environments is still a challenge. This can be mainly attributed to the lack of significant motion patterns and limited variations in the environment. In most of the office activities (like reading, copying, browsing, watching a video, writing ), the variability in image background, as observed from the egocentric video is limited. This yields poor accuracy due to the lack of sufficient discriminative information. However, a fusion of these features could improve the performance. The visual features can provide a context for the action, and the combination of ego-motion and eye movement pattern can result in better accuracy in the overall classification.

%

\begin{figure*}[t]
\begin{center}
\includegraphics[width=1\linewidth]{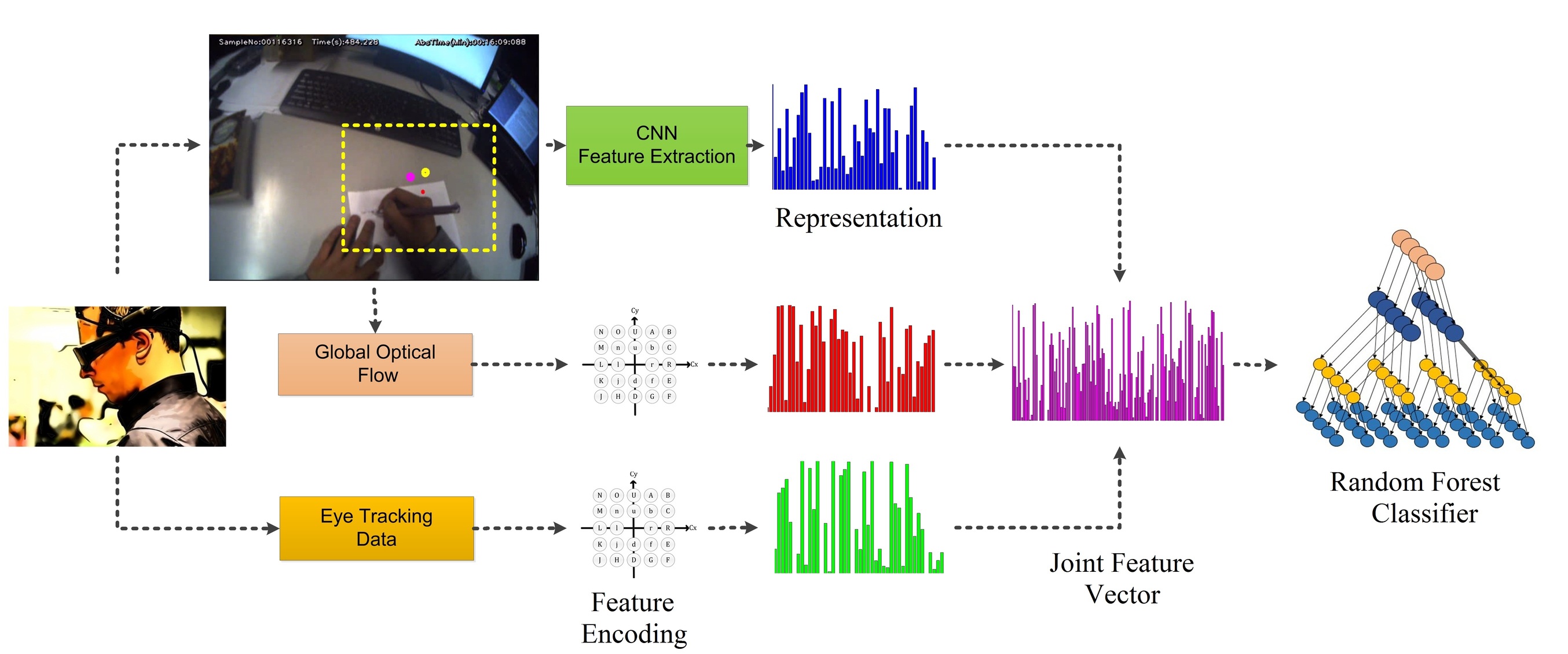}
\end{center}
\caption{The proposed framework, three channels of information are fused to classify the activities.}
\label{fig:framework}
\end{figure*}
\section{Proposed method}

In this work we propose to use information from the image, gaze locations and ego-motion for the recognition of activities. The features extracted from each domain along with the proposed fusion scheme is described below. A schematic diagram of the proposed framework is shown in Fig. \ref{fig:framework}.

\subsection{Feature extraction from image}

\begin{figure}[h]
\begin{center}
\includegraphics[width=1\linewidth]{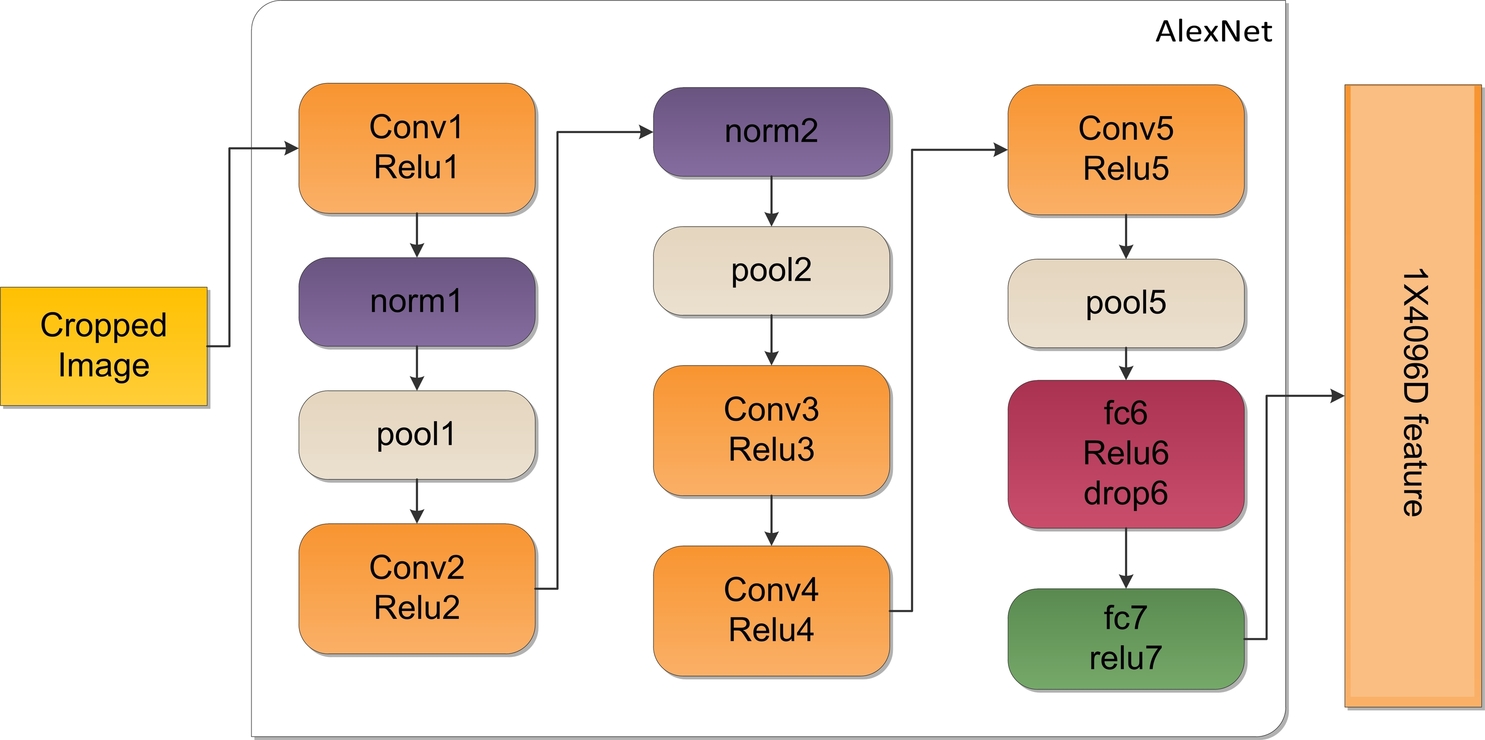}
\end{center}
\caption{CNN feature extraction scheme, cropped and resized image is fed into the pretrained network, outputs from \textit{fc7} are used as the feature.}
\label{fig:alex}
\end{figure}

Location of gaze on the images captured from a first person view (ego-centric) cameras carries valuable information which might be useful for activity classification. Previous works \cite{shiga2014daily} have used dense SIFT descriptor with PCA in a Bag of words (BoW) framework. Features were extracted from the patch around the point of gaze.  They computed the descriptors for each frame separately. The accuracy of this method could fall when the training and testing environments are different. For example, the appearance of a book might differ with variations in size, pose, color, and different types of binding. Ideally, the feature representation should be invariant to such changes as it is intended to give a context to the actions. We have used convolutional neural network \cite{sharif2014cnn} based feature extractor in this work owing to its high representation power. A pre-trained Alexnet model \cite{krizhevsky2012imagenet} (trained on the Imagenet dataset) is employed for this purpose. The fully connected output layer was removed, and a feature descriptor of dimension 4096 was obtained. The architecture of Alexnet excluding the final fully connected layer is shown in Fig. \ref{fig:alex}. We take the output from \textit{fc7} layer after applying the rectified linear unit (ReLU) transformation \cite{gong2014multi}.
For each image in the training set, a patch of size $200 \times 200$ was selected around the gaze location. The image patch obtained was resized and fed to the CNN to obtain a 4096-dimensional feature vector. We have extracted features from all the images in the training set in a similar manner. K-means clustering was performed on this data, and 15 cluster centers were kept. Now, for each image, the feature representation is computed, and the cluster center closest to it is found out. Histogram Voting across the cluster centers are carried out, and the normalized votes are computed in a temporal window of 25 seconds. The histogram obtained is used as the feature input for the activity classification.

\subsection{Feature extraction from eye tracking data}

The eye movement sequence is of the form
\begin{equation}
E = \left\{ {{e_{x,t}},{e_{y,t}}} \right\}_{t = 1}^{{T_E}}
\end{equation}

where,  ${e_{x,t}},{e_{y,t}}$ denote the $x$ and $y$ components of gaze position at the time instant $t$. $T_E$ denotes the duration of the sequence. The raw sequence is median filtered to remove noise.  Let ${e_{x,t}}$ be the input signal corresponding to the $x$ component of eye movement. The wavelet coefficient $Cx_b^a$
of ${e_{x,t}}$ at scale $a$ and position $b$ is defined as

\begin{equation}
Cx_b^a = \int\limits_\Re  {{e_{x,t}}{1 \over {\sqrt a }}} \overline {\psi \left( {{{t - b} \over a}} \right)} dt
\end{equation}

Continuous 1D wavelet coefficients are computed at a scale 10 using Haar-wavelet function.

Now, the wavelet coefficients are computed separately for $x$ and $y$ directions. The coefficients obtained are quantized as
\begin{equation}
 \hat Cx_b^a = \left\{
\begin{array}{ll}
      2 & {\tau _{large}} \le {Cx_b^a} \\
      1 & {\tau _{small}} < Cx_b^a \le {\tau _{large}} \\
      0 & { - {\tau _{small}} \ge Cx_b^a \le {\tau _{small}}} \\
      -1 & { - {\tau _{large}} < Cx_b^a \le  - {\tau _{small}}} \\
      -2 & {Cx_b^a \le  - {\tau _{large}}}  \\
\end{array}
  \right.
\end{equation}

where, $\tau _{large}$ and $\tau _{small}$ are empirically decided thresholds.

\begin{figure*}[h]
\begin{center}
\includegraphics[width=0.6\linewidth]{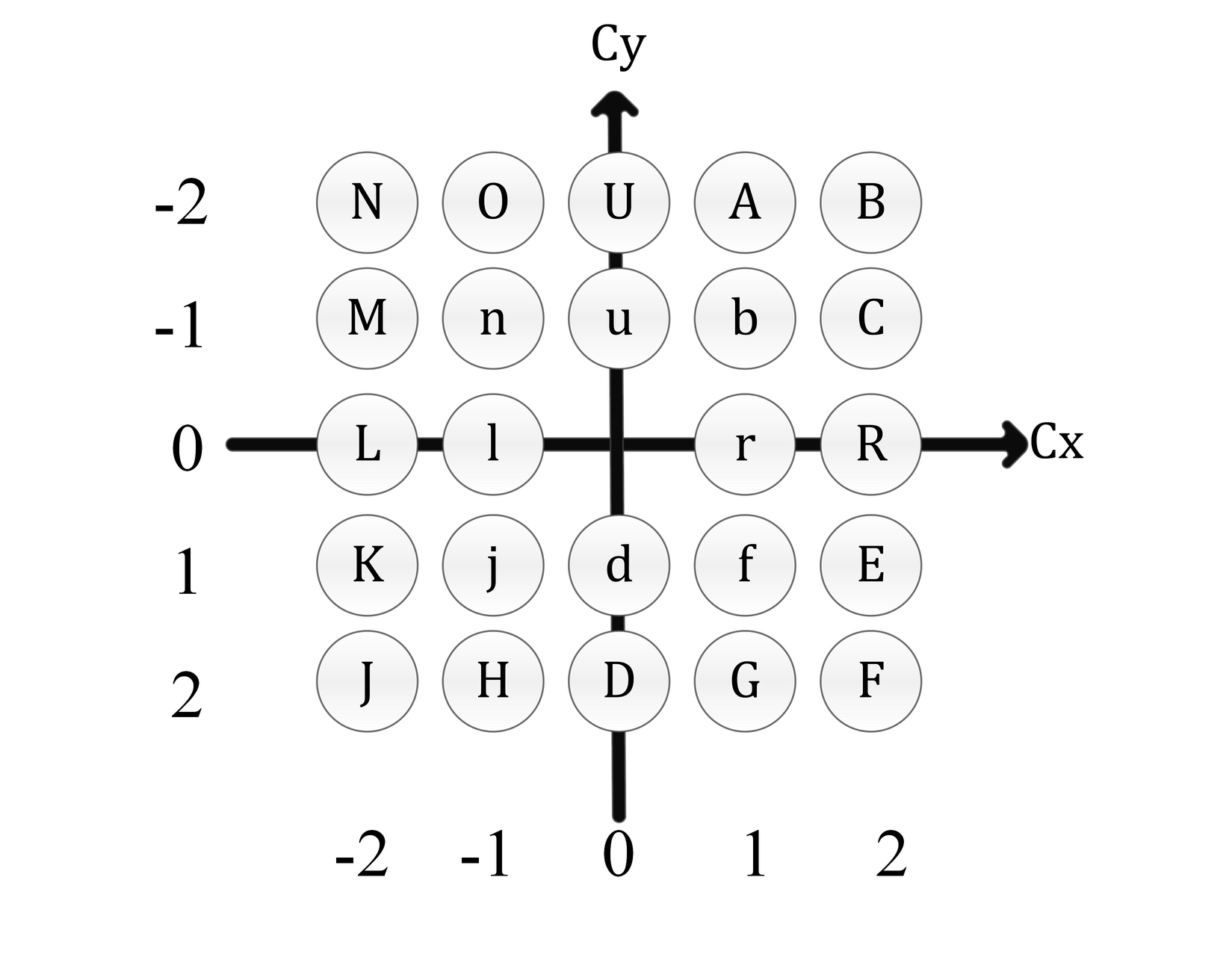}
\end{center}
\caption{Motion encoding scheme.}
\label{fig:motionencode}
\end{figure*}

$Cy_b^a$ is also quantized to $\hat Cy_b^a$ in a similar manner.

Based on the joint sequence $(\hat Cx_b^a,\hat Cy_b^a)$, a string sequence is generated as in Fig. \ref{fig:motionencode}.

The normalized histogram of the string sequence over a sliding temporal window is used as the feature for classification.

\subsection{Feature extraction from motion}

Motion features are extracted from the optical flow between subsequent frames.
Let the $i^{th}$ frame be denoted as $F_i$. For each frame, corner detection is performed to obtain the candidate points 
to track. The points are tracked using Lucas-Kanade optical flow. Successfully tracked points are found out using forward-backward error \cite{kalal2010forward}. The median flow between the frames can be computed as
\begin{equation}
\Delta x = median(\delta {x_j}),\,j \in [1,K]
\end{equation}
\begin{equation}
\Delta y = median(\delta {y_j}),\,j \in [1,K]
\end{equation}

Where $K$ is the number of sparse points tracked between $F_i$ and $F_{i+1}$, and $\delta {x_j}$ and $\delta {y_j}$ denote the optical flow of $j^{th}$ point in $x$ and $y$ direction respectively.

Once the global optical flow is obtained, we use a similar encoding scheme as used for eye gaze data. The histogram of the encoded sequence obtained over a temporal window is used as the feature for the classification task.

\subsection{Fusion and classification framework}

Features obtained from the three independent modalities namely ego-motion, eye gaze features and visual features are combined in the proposed approach. Feature level fusion \cite{ross2005feature} is adopted where three modalities are concatenated to form the final feature vector. We have extracted all the features using a temporal sliding window of 25 seconds with a stride of one second. Histogram of each independent feature is computed and concatenated for training the classifier model.

The classification model chosen should be able to handle different types of data as inputs. We have chosen Random Forest (RF) Classifier for this task. Random forest algorithm is an ensemble of decision trees initially proposed by Breiman \cite{breiman2001random}. It can intrinsically handle multi-class classification problems.  Instead of using a single tree for classification, predictions from a large number of trees are integrated to form the final prediction. Different trees in the forest are trained from bootstrap samples. The original data is sampled with replacement and trees are trained using these bootstrap samples. For each tree, a subset of predictors are randomly selected at each node and an optimal split is found \cite{ma2005classification}. The tree is grown without pruning.  In the testing phase, the test sample is fed to $N$ trees in the forest. Each tree makes a prediction by evaluating the decision tree. The final prediction is obtained using voting strategy among the outputs of $N$ decision trees. Random forest is robust to noise and faster to train. RF gives better predictions without overfitting due to the out of bag error cross-validation used during the training.  


\section{Experiments and results}

 Activities performed in office environments are considered in the experiments as they are difficult to classify by other methods. We have evaluated the accuracy of individual features as well as joint representation in a multi-class scenario to assess their performance.

\subsection{Database used}
We have used UTokyo First-Person Activity Recognition Dataset \cite{ogaki2012coupling} for the evaluation task.  The dataset contain the data recording of five subjects performing five different actions in an office environment. The classes available were reading a book, watching a video, copying text, writing on paper, and internet browsing.  Each of these activities was performed for two minutes. There was a time gap of thirty seconds (`Void' class) between each activity where subjects were allowed to converse, sing and move freely. Each subject performed the activities twice. The data from these two sessions were used as the training and test sets.  The recordings obtained from EMR-9 eye tracking device was also available with the dataset. For analysis purpose, we have used the eye tracking data and the low-resolution video ($640 \times 480$ resolution) from the dataset.

\subsection{Experiment protocol}
For each subject in the dataset, the features corresponding to visual, eye movement and ego-motion were extracted from the dataset.  For each subject, two separate instances of the same activity class are available. We have used these two folds for the evaluations. Initially, the first fold was used for training and the second one for testing. In the second fold, training and testing sets were interchanged and the average accuracy computed across these two sets are reported. The evaluations were performed for a multi-class scenario, data across all subjects were used for training and testing.


\subsection{Multi-class classification}
 We have analyzed the performance with two different scenarios namely five class and six class classification. In the latter, `Void' class is also used as a valid label.

\subsubsection{Experiments with five activity classes}

\begin{figure*}[!htb]
\begin{center}
\includegraphics[width=1\linewidth]{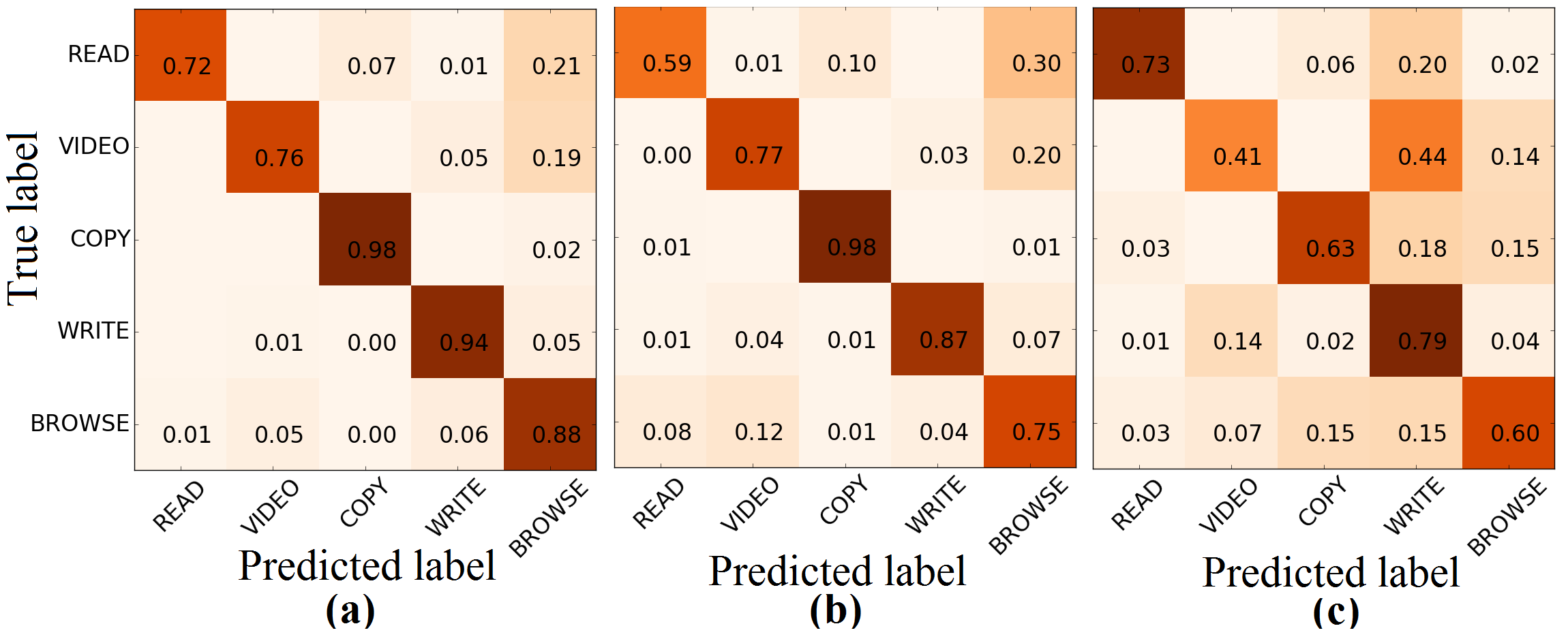}
\end{center}
\caption{Normalized confusion matrix for five classes, a) Combined features, b) Joint Ego-Eye motion feature, c) Visual features }
\label{fig:conf_fiveclass}
\end{figure*}

We have used five activity classes in this trial. Experiments were performed in multiclass classification scenario to evaluate the generalization capability of the features. Training and testing were done across all the individuals. The first session data from all the subjects were used for training. A Random Forest model was trained using the joint feature vector obtained from ego-motion, eye motion, and CNN features. The individual accuracy of the modalities was also tested by training separate models for CNN as well as joint eye-ego motion features.  The experiment was also performed by interchanging the training and testing sets. The average results among these two folds were found. The normalized confusion matrix obtained is shown in Fig. \ref{fig:conf_fiveclass}. The individual confusion matrices for visual and motion features alone are also shown in Fig. \ref{fig:conf_fiveclass}. The average accuracy over multiple runs is shown in Table. \ref{tab:avg_accuracy}.

\subsubsection{Experiments with six activity classes}

\begin{figure*}[!htb]
\begin{center}
\includegraphics[width=1\linewidth]{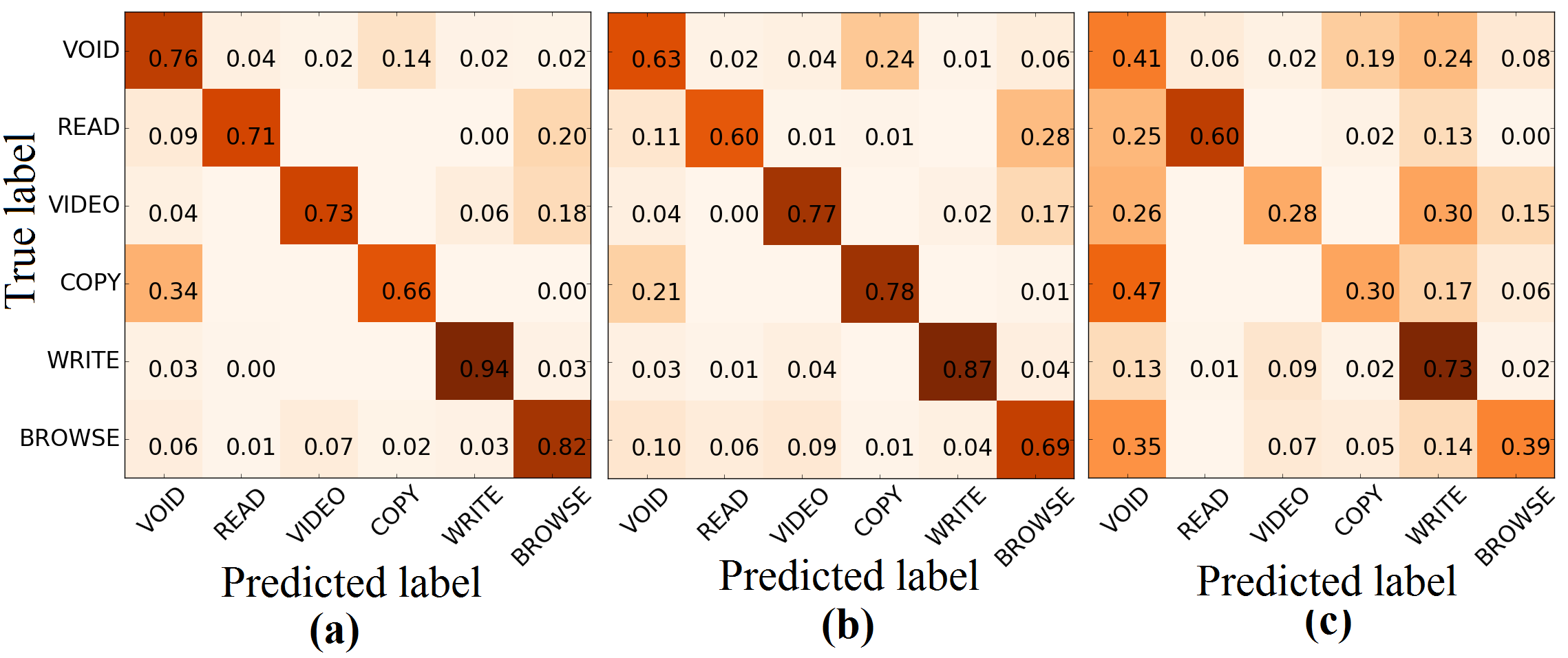}
\end{center}
\caption{Normalized confusion matrix for six classes, a) Combined features, b) Joint Ego-Eye motion feature, c) Visual features }
\label{fig:conf_sixclass}
\end{figure*}

In this experiment, we have considered all six classes including the 'Void' class. We have followed similar testing methodology as described for five class scenario. The results obtained are shown in Fig. \ref{fig:conf_sixclass} and Table \ref{tab:avg_accuracy}.

\subsubsection{Accuracy across different subjects}
\begin{figure}[h]
\begin{center}
\includegraphics[width=0.99\linewidth]{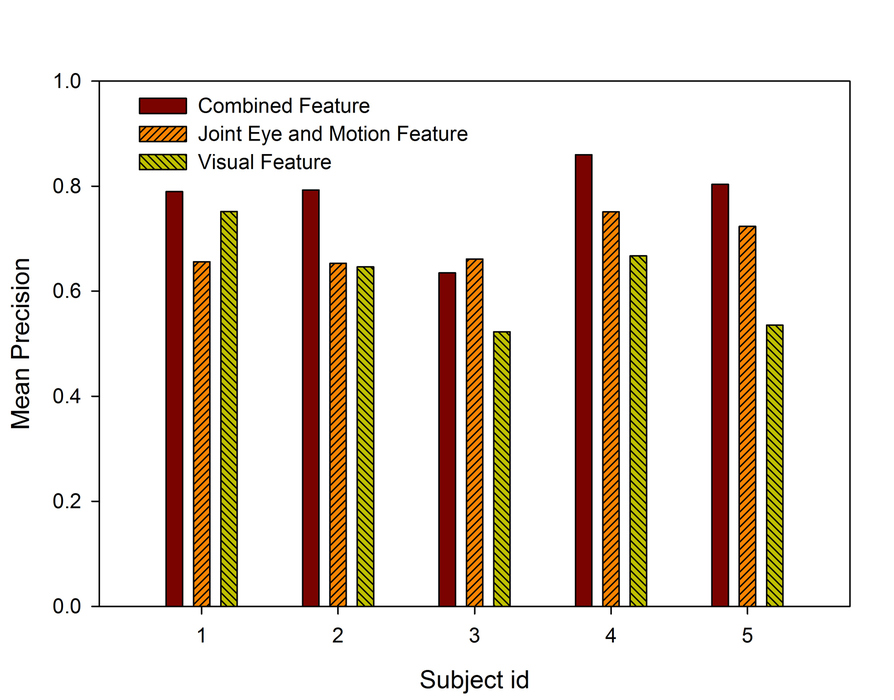}
\end{center}
\caption{Variation of accuracy across different subjects}
\label{fig:acc_subjects}
\end{figure}

The variations in accuracy across different subjects are shown in Fig. \ref{fig:acc_subjects}.
The combined feature gives better results for most of the subjects. 
\begin{table}[h]
\centering
\caption{Average accuracy for all three for both 5 and 6 class scenarios}
\label{tab:avg_accuracy}
\begin{tabularx}{1\linewidth}{@{\extracolsep{\fill}}llll@{}}
\toprule
Classes & \begin{tabular}[c]{@{}l@{}}Combined\\  Feature\end{tabular} & \begin{tabular}[c]{@{}l@{}}Eye and Ego Motion\\ Feature\end{tabular} & \begin{tabular}[c]{@{}l@{}}Visual  (CNN)\\ Feature\end{tabular} \\ \midrule
6 class & \textbf{77.09\%  }                                                   & 72.49\%                                                              & 45.03\%                                                         \\
5 class & \textbf{85.65\% }                                                    & 79.38\%                                                              & 62.97\%                                                         \\ \bottomrule
\end{tabularx}
\end{table}

\subsubsection{Accuracy across classes}

\begin{figure}[h]
\begin{center}
\includegraphics[width=0.99\linewidth]{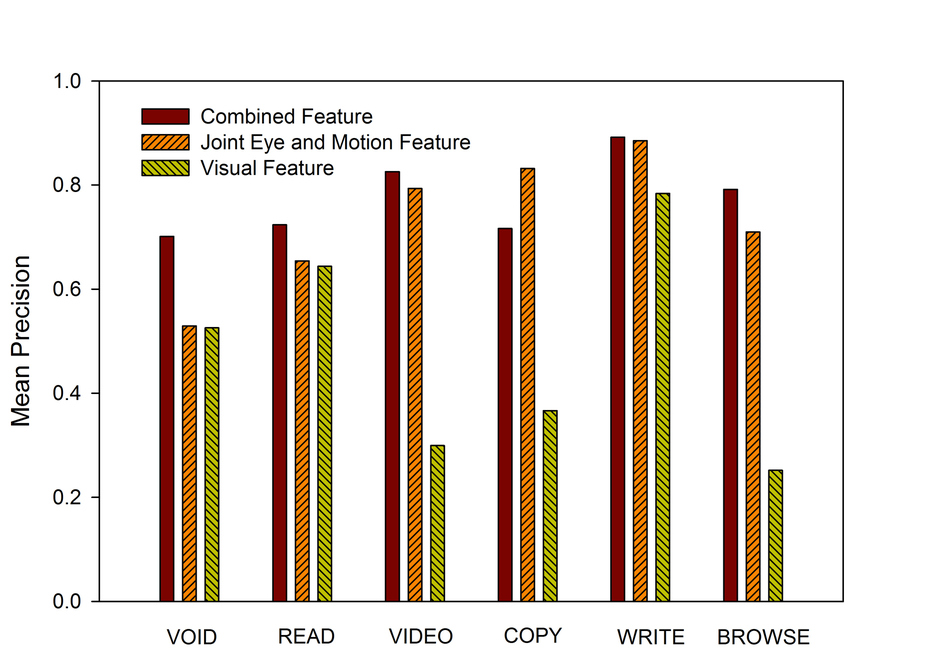}
\end{center}
\caption{Variation of accuracy across different classes}
\label{fig:acc_classes}
\end{figure}

The accuracy of different classes for different feature combinations are shown in Fig. \ref{fig:acc_classes}. The joint representation achieves better results as compared to the individual features. The joint eye-ego motion feature obtains the best accuracy among the features. The `Void' class shares similar visual features and motion features as the subjects were allowed to interact freely during those periods. This could explain the low accuracy of the `Void' class. Visual features give good results in activities like `Write' and `Read' since the field of view is different from other activities. 

\begin{figure}[h]
\begin{center}
\includegraphics[width=0.99\linewidth]{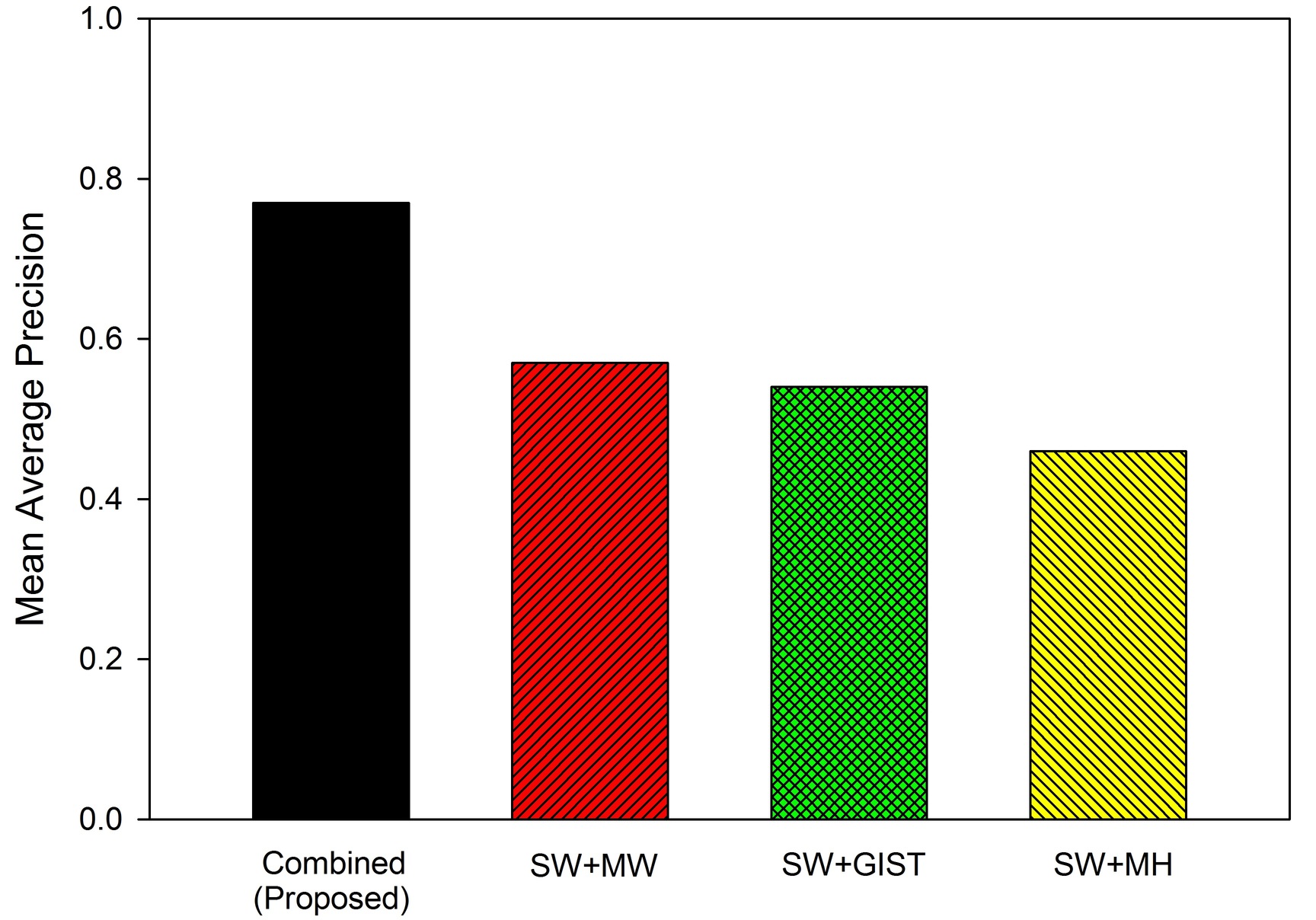}
\end{center}
\caption{Comparison with state of the art methods \cite{ogaki2012coupling}, SW+MW (Saccade Word+ Motion Word) \cite{ogaki2012coupling}, MH (Motion Histogram) \cite{kitani2011fast}, GIST \cite{oliva2001modeling}}
\label{fig:comparison}
\end{figure}

\subsection{Comparison with other methods}
We have compared the results obtained with different methods. Saccade word and motion word (SW + MW) \cite{ogaki2012coupling}, which is a combination of eye movement and egomotion `N-gram' features, obtains the second best result.
GIST features \cite{oliva2001modeling} extracts the visual content of the scene can be used for activity recognition in egocentric video \cite{spriggs2009temporal}. A combination of saccade word (SW) and GIST effectively combines motion and visual features. 
Motion histogram (MH) proposed by Kitani \textit{et al}.\cite{kitani2011fast} encodes the instantaneous as well as period motion using Fourier analysis. The accuracy of saccade word and motion histogram is also taken for comparison. The mean average precisions of the methods are compared in Fig. \ref{fig:comparison}. 

The proposed method outperforms all the other methods. The addition of
visual features along with the motion and eye gaze features improved the accuracy significantly. Compared to other methods, the higher representation power of the CNN based feature and the combination of ego-eye motion features makes the algorithm more accurate.

\subsection{Discussions}

 From the results obtained, it can be seen that the addition of three modalities improves the accuracy. In six class scenario, highest accuracy is achieved for class `Write'. This can be attributed to both distinct gaze patterns as well as visual features during the activity. Especially, the high accuracy of visual features during this activity may be due to the appearance of paper and pen which are unique to this activity. Even though the addition of visual features increases the overall accuracy, the individual performance of visual features in many cases are poor. The activities used in this experiment were performed in an office environment,which does not have much diversity in visual information. The addition of the `Void' class introduces more errors as the same visual features appear in multiple activities.

In the five class scenario, `Void' class was not present. The accuracy of visual features is much better than the six class case. The overall accuracy of classification is also much better in this scenario.  The random forest based classifier tries to identify the important features for activity classification from the joint feature representation.

Some of the advantages of the proposed system are described here. Three distinct channels of information are fused in the proposed approach. This improves the generalizability of the approach for a larger number of classes. Representation
of one particular activity might not require the features from all three channels. For example, reading has a characteristic pattern as observed from eye tracking data (sequence of small fixations and saccades), It may be possible to identify reading activity from eye tracking data alone. Classifying browsing activity from watching movies might require all three channels of information. The high-level CNN descriptors used are suitable for giving a context to the actions. The random forest algorithm is capable of identifying important features which are relevant for the identification of a particular action. The framework can determine the important features required for classifying the activities accurately. Even though the activity classes used in this work are small, the framework is capable of handling a large number of classes. The Random Forest based classifier can compute the features relevant for identifying each action.

\section{Conclusions}

In this work, we have proposed an approach for combining different modalities such as ego-motion features, eye movement features and visual features for classification of activities. A joint feature vector is formed from the individual feature extractors, and a random forest classifier was used to classify the activities using this joint representation.  Joint eye-ego motion feature gave the best individual accuracy among the features. However, the addition of visual feature resulted in a higher accuracy in activity classification. Additional channels of information can be easily added to the framework. The addition of activity-dependent object detectors and a weighted fusion of these three modalities might improve the results.

\bibliographystyle{IEEEtran}

\bibliography{refs_gazeactivity}

\end{document}